\pdfoutput=1
\documentclass{article}

\usepackage[T1]{fontenc}
\usepackage{arxiv_preprint,times}
\usepackage[scaled=0.88]{helvet}
\usepackage{microtype}
\usepackage{amsmath,amssymb}
\usepackage{booktabs}
\usepackage{graphicx}
\usepackage{xcolor}
\usepackage{tikz}
\usetikzlibrary{positioning,arrows.meta,calc}
\usepackage{hyperref}
\usepackage{url}
\usepackage{xurl}
\usepackage{capt-of}
\newcommand{\modelfont}[1]{\textsf{#1}}
\newcommand{\methodname}{\modelfont{C2R\#}}
\newcommand{\method}{\methodname{} model}
\newcommand{\dinorname}{\modelfont{DINO-R2C}}
\newcommand{\dinor}{\dinorname{} model}
\newcommand{\vone}{\modelfont{C2R orig.}}
\newcommand{\vonepaper}{original \modelfont{C2R} paper}
\newcommand{\framework}{pair-free C2R}
\newcommand{\cadj}{coarse-to-real}

\newcommand{\radj}{real-to-coarse}
\newcommand{\pcoarse}{pseudo-coarse}
\newcommand{\Pcoarse}{Pseudo-Coarse}
\newcommand{\modGeomCoarser}{Coarser geometry}
\newcommand{\modGeomCoarse}{Coarse geometry}
\newcommand{\modAnimeCoarser}{Coarser Anime}
\newcommand{\modAnimeCoarse}{Coarse Anime}
\newcommand{\gameRenders}{game renders}
\newcommand{\geometryBlockouts}{geometry blockouts}
\newcommand{\stylizedAnimations}{stylized animations}
\newcommand{\projecturl}{\url{https://gonzalognogales.github.io/coarse2real-sharp/}}

\title{Pair-Free Coarse-to-Real Video Generation via Real-to-Coarse Image Supervision}

\author{
Gonzalo G\'omez-Nogales$^{1,\dagger}$ \quad
Daohan Lu$^{2,3}$ \quad
Peiye Zhuang$^{3}$ \quad
Tinghui Zhou$^{3}$ \quad
Yi Zhou$^{4,\dagger}$\\
{\normalfont $^{1}$Universidad Rey Juan Carlos \quad
$^{2}$New York University \quad
$^{3}$Roblox \quad
$^{4}$Adobe}
}

\hypersetup{
  pdftitle={Pair-Free Coarse-to-Real Video Generation via Real-to-Coarse Image Supervision},
  pdfauthor={Gonzalo G\'omez-Nogales, Daohan Lu, Peiye Zhuang, Tinghui Zhou, Yi Zhou}
}

\date{}

\begin{document}
\maketitle
{\renewcommand{\thefootnote}{\fnsymbol{footnote}}\footnotetext[2]{Work done while at Roblox.}}

\begin{abstract}
Coarse-to-real (C2R) video generation converts coarse animation, simulation, or
game-rendered videos into photorealistic videos while preserving scene layout,
camera motion, and dynamics. Training such models requires aligned (coarse,
realistic) video pairs, which are expensive to create and confine prior
systems to narrow synthetic domains. We exploit a simple asymmetry: paired
videos are hard to obtain, but paired \radj{} images are easy to construct
with existing editing and 3D reconstruction tools. Using these image pairs, we
train the \dinor{} to transfer realistic images to coarse images in DINO
feature space, and apply it frame-wise to ordinary real videos to
synthesize pseudo-coarse videos. The \method{} then learns to generate each
real video from its pseudo-coarse counterpart, free from needing (coarse,
realistic) video pairs. At inference, the \dinor{} is discarded and actual
coarse videos drive the \method{} directly. Thus, our pair-free training scales with any
real-video corpora rather than with paired synthetic videos.
We will release the model and project webpage at \projecturl{}.
\end{abstract}

\section{Introduction}

\begin{figure}[!htbp]
\centering
\includegraphics[width=\linewidth]{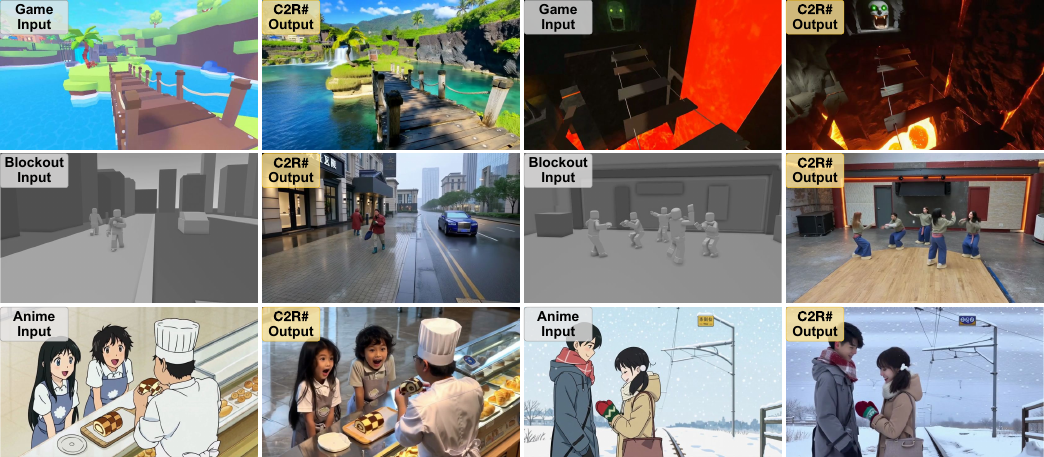}
\caption{The \method{} transforms \gameRenders{} (top), \geometryBlockouts{} (middle),
and \stylizedAnimations{} (bottom) into photorealistic videos.}
\label{fig:teaser}
\end{figure}

Coarse-to-real (C2R) video generation uses coarse videos, such as
\gameRenders{}, \geometryBlockouts{}, or \stylizedAnimations{}, as the structural
control for photorealistic video generation. The coarse input specifies scene
layout, camera motion, and dynamics, while a modern video model produces the
final visual appearance.

A major limitation of existing \cadj{} systems is that they need paired
training data. The \vonepaper{}~\citep{gomeznogales2026coarsetoreal}
introduces a model that learns controllability from paired synthetic (coarse, realistic)
videos. Such aligned video pairs are expensive to create, and the cost
multiplies with every additional coarse modality, so paired-video supervision is
difficult to scale.\looseness=-1

Our key observation is that this bottleneck is asymmetric. Paired \cadj{}
videos are expensive, but paired \radj{} images are cheap. Modern image editing
models and 3D fitting methods can turn diverse realistic images into aligned
coarse counterparts. Image-level supervision is therefore highly accessible.
However, using an image generator such as GPT-Image 2~\citep{openai2026gptimage2}
to edit every frame of a large video corpus into its coarse counterpart would
incur substantial generation costs.\looseness=-1

We exploit this asymmetry with an image-level \radj{} bridge, the \dinor{}.
Trained on (coarse, realistic) image pairs, it predicts the DINO
features~\citep{simeoni2025dinov3} of the coarse counterpart of a realistic
image. Applied to every frame of an ordinary real video, it synthesizes a
pseudo-coarse video that is paired with its source. We train the
\method{} to generate the real video from this pseudo-coarse control video. At
inference, the \dinor{} is discarded and an actual coarse video drives the
\method{} directly.

Using the image-level bridge poses two challenges. First, the \dinor{} is an image model, so its per-frame
outputs are not temporally smooth. As a result, we apply camera augmentation during its training
to reduce this jitter, and the \method{} learns to tolerate the rest because its
target is always the temporally coherent real video. Second, the predicted
features can leak cues of the realistic input that no actual coarse video
contains. Therefore, we block the leak by decoding the prediction to pixels with a
frozen RAE decoder~\citep{zheng2026diffusion} and re-encoding those pixels
before they serve as control.

This formulation moves the expensive part of supervision from videos to
images. Once the \dinor{} is trained for a coarse modality, any real-video
collection becomes \cadj{} training data, while supporting a new modality
requires only new image pairs rather than a new paired-video dataset.\looseness=-1

\section{Related Work}

\paragraph{Generative rendering from coarse structure.}
Turning cheap renders into realistic imagery predates diffusion models:
\citet{richter2021enhancing} enhance game frames with a G-buffer-conditioned
network trained adversarially against unpaired photographs, and
vid2vid~\citep{wang2018vid2vid} learns paired label-to-video synthesis. Recent
work conditions image or video diffusion models on G-buffers
\citep{beisswenger2025framediffuser,zeng2026rgbxnext}, animated
meshes~\citep{cai2024generativerendering}, physics-engine
states~\citep{lin2026alayarenderer}, or multimodal spatial
maps~\citep{alhaija2025cosmostransfer}, and RealMaster~\citep{cohenbar2026realmaster}
lifts rendered videos to photorealism by propagating enhanced keyframes to build
its own paired videos. NVIDIA's DLSS 5~\citep{nvidia2026dlss5} conditions a
real-time generative renderer on rendered frames and engine motion vectors
to enhance photorealistic appearance. All of these obtain supervision from inside a rendering
pipeline, from per-scene synthesis, or from authored video pairs. We assume only
a coarse RGB video at test time and no paired videos at all.

\paragraph{The \vonepaper{}.}
Our direct predecessor~\citep{gomeznogales2026coarsetoreal} conditions a Wan2.1
backbone on DINOv3 features of the control video through a lightweight adapter.
Its control pathway is trained mostly on unpaired real videos reconstructed
from their own DINO features, with a small set of paired synthetic (coarse,
realistic) videos anchoring controllability. Controllability therefore rests
on scarce, domain-limited video pairs, and the adapter sees mostly features of
real videos in training but features of coarse videos at test time. Our
\framework{} framework keeps the objective and control interface but removes
paired videos from training.

\paragraph{Controllable video diffusion.}
Video diffusion and flow-matching models can be guided by depth, edges,
segmentation, or sparse keyframes through ControlNet-style
branches~\citep{zhang2023adding,esser2023gen1,guo2024sparsectrl,lin2025ctrladapter},
by unified context blocks~\citep{jiang2025vace,wan2025wan}, or by warped
noise that carries motion~\citep{burgert2025gowiththeflow}. For image generation,
LooseControl~\citep{bhat2024loosecontrol} generalizes ControlNet's depth
conditioning to loosely specified scene boundaries and 3D object boxes.
Instruction-based
video editors such as Editto~\citep{bai2025editto} are trained on large
synthetic edit datasets and can perform sim-to-real edits, which is why we
compare against them. Explicit structural controls preserve geometry strongly
but can over-constrain generation or leak the appearance of the control
input; as in the \vonepaper{}, the \method{} conditions on implicit DINO
features, which carry layout and semantics while leaving appearance to the
video model.

\paragraph{Self-supervised features as a cross-domain bridge.}
DINO features~\citep{oquab2024dinov2,simeoni2025dinov3} are dense and robust
to appearance change, and several works exploit this for domain transfer.
Concurrently, Control-DINO~\citep{dominici2026controldino} conditions an
image-to-video model on DINOv3 features for stylization and video-from-3D,
and \citet{liu2026ssb} use self-supervised features as an appearance-invariant
bridge for unpaired image translation. Earlier unpaired video translation
relied on cycle consistency~\citep{zhu2017cyclegan} with pseudo-supervision
from synthetic motion~\citep{wang2022pseudoflow}, while
InstructPix2Pix~\citep{brooks2023instructpix2pix} and Ditto~\citep{bai2025editto}
showed that strong image models can synthesize the paired data an editor
needs. We combine both ideas: image models provide cheap (coarse, realistic)
image pairs, an \radj{} model learned in DINO space and decoded through an RAE
decoder~\citep{zheng2026diffusion} lifts them to videos, and the resulting
pseudo-paired videos supervise the \cadj{} generator.

\section{Method}
\label{sec:method}

\begin{figure}[t]
\centering
\includegraphics[width=\linewidth]{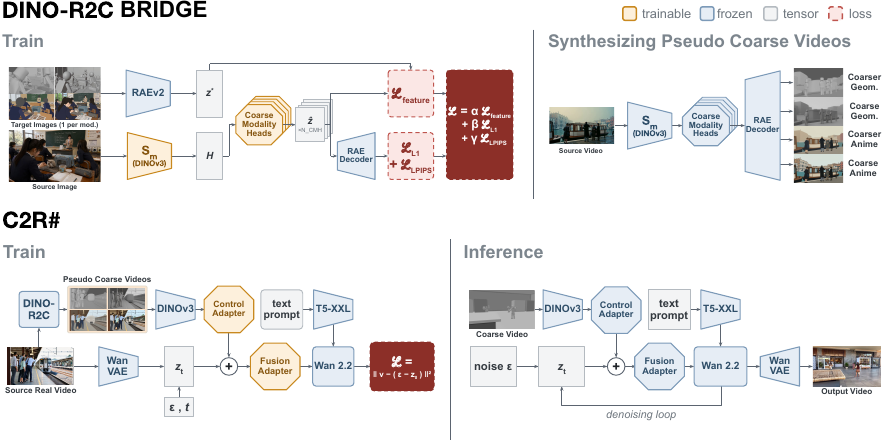}
\caption{\textbf{Overview of the \dinorname{} and \methodname{} models.}
\textbf{Top:} The \dinor{} is trained from paired realistic and coarse images using
feature reconstruction and pixel-space reconstruction losses. Once trained, it
is applied to real videos frame-wise to synthesize pseudo-coarse videos.
\textbf{Bottom left:} The \method{} is trained on these pseudo-paired videos.
The pseudo-coarse video is encoded by the frozen DINO encoder $E$ and injected into the frozen
Wan2.2 backbone through trainable control and fusion adapters, while the original
real video provides the training target.
\textbf{Bottom right:} at inference, an actual coarse video is encoded directly
by $E$ and used to condition the same \method{}; the \dinor{} is no longer
required. Orange modules are trainable, blue modules are frozen, gray boxes are
tensors, and red boxes denote training losses.}
\label{fig:full-pipeline}
\end{figure}

\subsection{Problem Setup}
\label{sec:setup}

At inference time, the user provides an actual coarse video $v^{\mathrm{c}}$
from categories such as \gameRenders{}, \geometryBlockouts{}, or \stylizedAnimations{}, together
with a text prompt. We seek a generator that produces a realistic video while
preserving the scene layout, camera motion, and dynamics of $v^{\mathrm{c}}$.
For training, we assume access to ordinary real videos $v^{\mathrm{r}}$ and to
aligned (coarse, realistic) \emph{image} pairs, but to no paired
$(v^{\mathrm{c}}, v^{\mathrm{r}})$ videos. Throughout, superscripts
$\mathrm{r}$ and $\mathrm{c}$ denote realistic and coarse data.\looseness=-1

Our \framework{} framework has two models (Figure~\ref{fig:full-pipeline}).
The \dinor{} (Section~\ref{sec:dinor2c}) learns \radj{} image transfer from
the image pairs and is then applied to every frame of the real videos,
producing pseudo-paired videos (Section~\ref{sec:pseudo-pairs}). The \method{}
(Section~\ref{sec:c2r}) is trained on these pseudo-paired videos to generate
the real video from its pseudo-coarse counterpart; at inference it is driven
directly by actual coarse videos.

\subsection{Image-Level Real-to-Coarse Bridge: the \dinorname{}}
\label{sec:dinor2c}

\begin{figure}[t]
\centering
\includegraphics[width=\linewidth]{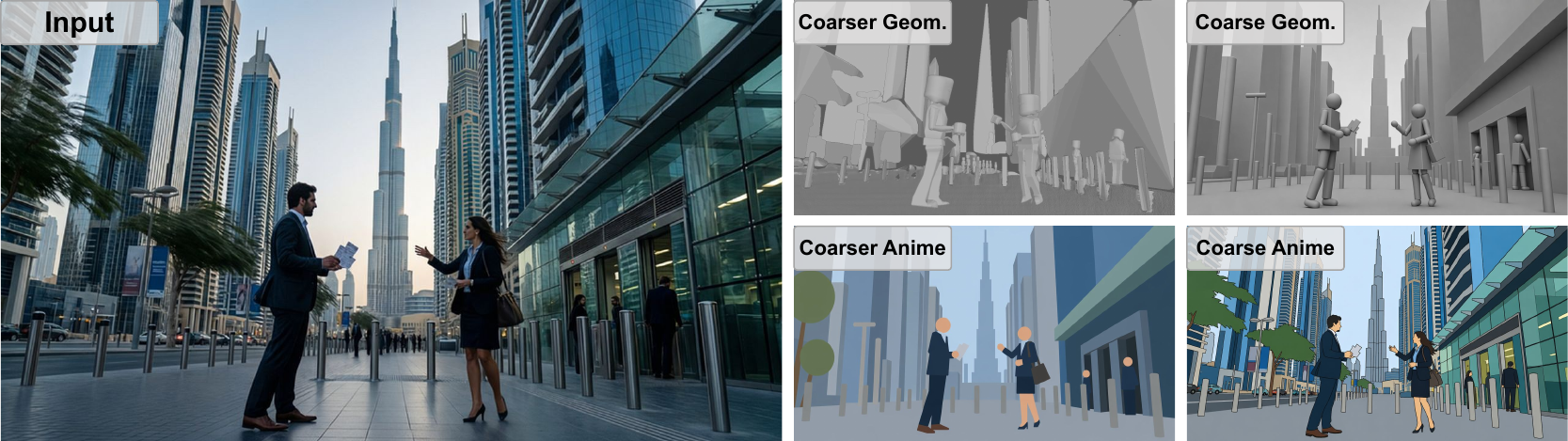}
\caption{Example (coarse, realistic) image pairs. A realistic image (left) is paired with coarse
images in four modalities (right):
\modGeomCoarser{}, \modGeomCoarse{},
\modAnimeCoarser{}, and \modAnimeCoarse{}.\looseness=-1
}
\label{fig:r2c-pairs}
\end{figure}

\paragraph{Image pairs.}
We define a \emph{coarse modality} as one visual form of coarse input that the
\method{} should support, and use the four modalities shown in
Figure~\ref{fig:r2c-pairs}. Each coarse image preserves the subjects, layout,
pose, and camera of its realistic counterpart. We generate realistic source
images with GPT-Image 2~\citep{openai2026gptimage2} and create their coarse
counterparts through instruction-guided editing with the same model or 3D
reconstruction and simplification. All images are produced
at $1280\times720$, with 14,000 aligned (coarse, realistic) image pairs per
modality (Appendix~\ref{app:pairs}). These image pairs are the only paired
supervision in our framework.

\paragraph{Model and objective.}
Let $E$ be a frozen DINO encoder (DINOv3-L/16~\citep{simeoni2025dinov3}) that
maps an image to a dense feature grid. For each coarse modality $m$, the
\dinor{} learns a mapping $S_m$ from a realistic image to the DINO features of
its coarse counterpart,
\begin{equation}
    S_m(I^{\mathrm{r}}) \approx E(I^{\mathrm{c}}_m),
    \qquad
    S_m = H_m \circ B,
\end{equation}
where $B$ is a trainable DINOv3-L backbone shared across all four modalities
and $H_m$ is one of four modality-specific residual MLP heads. We learn the
mapping in DINO feature space rather than in
RGB because the self-supervised representation captures scene structure and
layout while being relatively insensitive to the large appearance gap between
realistic and coarse images; empirically, training also converges more easily
in DINO space. To supervise the prediction in pixel space as well, a frozen RAE
decoder $D$~\citep{zheng2026diffusion} decodes it to a coarse image
$\hat{I}^{\mathrm{c}}_m = D(S_m(I^{\mathrm{r}}))$. The training objective is
\begin{equation}
    \mathcal{L}_{\mathrm{R2C}}
    =
    \lambda_{\mathrm{feat}}
    \big\| S_m(I^{\mathrm{r}}) - E(I^{\mathrm{c}}_m) \big\|_2^2
    +
    \lambda_{\mathrm{rgb}}
    \big\| \hat{I}^{\mathrm{c}}_m - I^{\mathrm{c}}_m \big\|_1
    +
    \lambda_{\mathrm{LPIPS}}\,
    \mathrm{LPIPS}\big(\hat{I}^{\mathrm{c}}_m, I^{\mathrm{c}}_m\big),
\end{equation}
a DINO feature reconstruction loss, an RGB reconstruction loss, and the LPIPS
perceptual loss~\citep{zhang2018unreasonable}, with
$\lambda_{\mathrm{feat}}=1.0$, $\lambda_{\mathrm{rgb}}=1.1$, and
$\lambda_{\mathrm{LPIPS}}=0.82$.

\paragraph{Camera augmentation.}
The \dinor{} is trained on image pairs but applied to every frame of a video,
where consecutive frames differ by small viewpoint changes. To close this
image-to-video gap, we apply the same random camera homography (crop,
translation, rotation, shear, and perspective) to a realistic image and its
coarse counterpart during training, so that the learned mapping is stable under
the viewpoint changes seen between frames. Section~\ref{sec:exp-bridge} ablates
the augmentation strength, and Appendix~\ref{app:augmentation} gives the exact
bounds.

\subsection{Constructing Pseudo-Paired Videos}
\label{sec:pseudo-pairs}

Applying the \dinor{} to every frame of a real video $v^{\mathrm{r}}$ yields a
pseudo-coarse video that remains paired with $v^{\mathrm{r}}$. Because each
frame is synthesized independently, the pseudo-coarse video carries some
temporal jitter. We use it only as the control video and always use the
temporally coherent real video as the target, which teaches the \method{} to
tolerate minor jitter in the control video while generating temporally
consistent output.

\paragraph{Decoding to RGB and re-encoding.}
The features $S_m(v^{\mathrm{r}})$ predicted by the \dinor{} already live in
DINO space and could condition the \method{} directly. However, they retain
hidden cues of the realistic input that are absent when an actual coarse video
is encoded at inference, a leakage that creates a train/test mismatch
(Section~\ref{sec:exp-c2r}). We therefore decode the prediction to coarse
pixels with the frozen RAE decoder, which discards information that cannot be
expressed as a coarse image, and re-encode those pixels with the same frozen
encoder $E$ used at inference. The training control features are thus
\begin{equation}
    C_{\mathrm{train}} = E\big(D(S_m(v^{\mathrm{r}}))\big),
\end{equation}
computed per frame and on the fly during \methodname{} training.

\subsection{Coarse-to-Real Video Generation: the \methodname{}}
\label{sec:c2r}

\paragraph{Architecture.}
The \method{} builds on a frozen pretrained text-to-video diffusion backbone,
Wan2.2 T2V-A14B~\citep{wan2025wan}, whose VAE, text encoder, and denoiser all
remain frozen. The denoiser is a mixture of two experts routed by diffusion
timestep, so we train one control adapter per expert, active only when its
expert is. Each control adapter takes the per-frame DINO control features $C$,
projects them to an intermediate width, applies residual space-time convolution
blocks with timestep conditioning, compresses time to the VAE latent frame
rate, and maps the result to the denoiser width through a zero-initialized
output head. At our operating resolution the DINO feature grid aligns
one-to-one with the backbone's spatial token grid, so no learned spatial
upsampling is needed. A zero-initialized fusion adapter then adds the control
features after the frozen patch embedding, so at initialization the model is
identical to the pretrained backbone. Architectural and training details are
given in Appendix~\ref{app:c2r-training}.

\paragraph{Training objective.}
For a target real video $v^{\mathrm{r}}$ with latent
$z_0 = \mathrm{VAE}(v^{\mathrm{r}})$, we sample a flow timestep $\sigma$ and
Gaussian noise $\epsilon$, form $z_\sigma = (1-\sigma) z_0 + \sigma \epsilon$,
and minimize the rectified-flow velocity loss~\citep{liu2023rectifiedflow}
\begin{equation}
    \mathcal{L}_{\mathrm{FM}}
    =
    \big\|
    G\big(z_\sigma, \sigma, e_{\mathrm{text}};\, A(C_{\mathrm{train}}, \sigma)\big)
    - (\epsilon - z_0)
    \big\|_2^2 ,
\end{equation}
where $G$ is the frozen denoiser, $e_{\mathrm{text}}$ is the text embedding,
and $A$ denotes the trainable adapters, the only parameters that are updated.
We use no control dropout: with a frozen backbone, steps without control
provide no gradient to the adapters.

\paragraph{Inference.}
The user provides an actual coarse video $v^{\mathrm{c}}$ and a text prompt.
The frozen encoder produces $C_{\mathrm{test}} = E(v^{\mathrm{c}})$, which
conditions the same adapters and backbone used in training; generation starts
from Gaussian noise, and the frozen VAE decodes the result. Neither the
\dinor{} nor the RAE decoder is used at inference.

\section{Experiments}
\label{sec:experiments}

We evaluate the \dinor{} (Section~\ref{sec:exp-bridge}) and the \method{}
(Sections~\ref{sec:exp-c2r} and~\ref{sec:exp-compare}). Because real-world coarse
videos have no photorealistic ground truth, the \method{} is evaluated with
reference-free quality, control-alignment, and VLM-judged realism metrics on
three evaluation sets of 100 actual coarse videos each: \gameRenders{},
\geometryBlockouts{}, and \stylizedAnimations{}. We report aggregate results
across all 300 videos. Dataset construction and metric definitions are given
in Appendices~\ref{app:bundles} and~\ref{app:metrics}.

\subsection{Training data.}
We train a shared four-modality \dinor{} on 14,000 image pairs per modality:
coarser geometry, coarse geometry, coarser anime, and coarse anime.
GPT-Image 2~\citep{openai2026gptimage2} generates the realistic images and edits
them into coarse geometry and both anime modalities; coarser geometry targets
are rendered from reconstructed and simplified 3D scenes.
Images are generated at $1280\times720$ using the procedure in
Appendix~\ref{app:pairs}; larger training sets yielded only marginal improvements.
For \methodname{} training, the \dinor{} synthesizes \pcoarse{} controls on
the fly for 389,854 real videos from DL3DV~\citep{ling2024dl3dv},
\texttt{SEKAI\_HQ}, and \texttt{SEKAI\_DRONE}~\citep{li2026sekai}.
The adapted VACE 2.2~\citep{jiang2025vace} and Editto
Sim2Real~\citep{bai2025editto} baselines use exactly the same real videos and
\pcoarse{} controls (Appendices~\ref{app:pseudo-pairs} and~\ref{app:baselines}).

\subsection{\dinorname{}}
\label{sec:exp-bridge}

\paragraph{Protocol.}
We evaluate each \dinorname{} model on 1,000 held-out images and 1,000 real
videos. Because each coarse modality has a different target distribution,
values are not comparable across modalities. All directly comparable ablations
target the \modGeomCoarse{} modality and vary the camera-augmentation strength
(none, $0.5\times$, $1\times$, $2\times$) and the inclusion of LPIPS
(Table~\ref{tab:r2c-video-ablation}; image ablations are in
Appendix~\ref{app:r2c-ablations}, Table~\ref{tab:r2c-image-ablation}).
Image metrics compare predictions with the paired coarse target. Video
metrics assess the stability of the \pcoarse{} video produced by applying the
model to each frame of a real video.

\paragraph{Video stability metrics.}
We estimate motion with RAFT-Small~\citep{teed2020raft}.
Perceptual flicker measures motion-aligned appearance changes using
LPIPS~\citep{zhang2018unreasonable};
smoothed-frame flicker measures broad color and shape changes using L1 on
blurred frames. Both are divided by the corresponding source-video error,
summarized by the median within each video, and averaged across videos.
Motion error measures disagreement between estimated source and output motion
vectors, while motion magnitude ratio indicates whether the output moves too
little or too much. Boundary stability measures the overlap of motion-aligned
outlines, and content consistency measures similarity between consecutive
DINOv3 frame features~\citep{simeoni2025dinov3}.
Appendix~\ref{app:r2c-video-metrics} explains all six
metrics and their interpretation.

\begin{table}[t]
\centering
\caption{\textbf{\dinorname{} design ablations, video stability.}
Evaluated on 1,000 real videos in the \modGeomCoarse{} modality at
$256\times256$. Columns report means across videos; both flicker measures are
per-video medians relative to source-video error. Bold/underlined values are
best/second-best. Metrics are explained in Section~\ref{sec:exp-bridge} and
Appendix~\ref{app:r2c-video-metrics}.}
\label{tab:r2c-video-ablation}

\begingroup
\setlength{\tabcolsep}{4pt}
\renewcommand{\arraystretch}{1.15}
\resizebox{\linewidth}{!}{%
\begin{tabular}{@{}lcccccc@{}}
\toprule
\dinorname{} Recipe
& \shortstack{Perceptual\\flicker\\(relative) $\downarrow$}
& \shortstack{Smoothed-frame\\flicker\\(relative) $\downarrow$}
& \shortstack{Motion\\error\\(EPE) $\downarrow$}
& \shortstack{Motion\\magnitude ratio\\$\approx 1$}
& \shortstack{Boundary\\stability\\(Dice) $\uparrow$}
& \shortstack{Content\\consistency\\(DINO) $\uparrow$} \\
\midrule
None, no LPIPS
& 2.2028 & 4.1565 & 2.2685 & 2.1909
& 0.7195 & 0.9816 \\

None + LPIPS
& 2.1793 & 3.9131 & 1.8678 & 1.8304
& 0.7271 & 0.9877 \\

$0.5\times$ + LPIPS
& 0.9314 & 1.7901 & 0.8819 & \underline{1.0844}
& 0.8345 & 0.9946 \\

$1\times$, no LPIPS
& \textbf{0.7965} & \underline{1.6825}
& 0.9066 & 1.1504
& 0.8327 & 0.9914 \\

$1\times$ + LPIPS
& 0.9023 & 1.7108
& \underline{0.8695} & 1.0875
& \underline{0.8354} & \underline{0.9946} \\

\addlinespace[2pt]
\textbf{$2\times$ + LPIPS (ours)}
& \underline{0.8649} & \textbf{1.6294}
& \textbf{0.8429} & \textbf{1.0797}
& \textbf{0.8376} & \textbf{0.9948} \\
\bottomrule
\end{tabular}%
}
\endgroup
\end{table}

\paragraph{Results.}
In Table~\ref{tab:r2c-video-ablation}, the selected recipe achieves the
lowest smoothed-frame flicker and motion error, highest boundary
stability and content consistency, and motion magnitude ratio closest to one.
It ranks second in perceptual flicker.
Its motion magnitude ratio of $1.0797$ indicates approximately
$8\%$ higher estimated motion magnitude than the source;
combined with the lowest motion error, this supports stronger
source-motion agreement among the evaluated variants.

With LPIPS enabled, increasing augmentation from
$1\times$ to $2\times$ improves all six reported video
metrics. At fixed $1\times$ augmentation, adding LPIPS
improves image perceptual reconstruction, video motion
agreement, motion-magnitude preservation, boundary stability,
and content consistency, but increases both relative flicker errors.
These results support selecting $2\times$ augmentation
with LPIPS as a balance between paired reconstruction
fidelity and temporal behavior during per-frame video
inference, rather than as the optimum for every
individual metric. Image reconstruction results and their tradeoffs are
reported in Appendix~\ref{app:r2c-ablations}.

\paragraph{Qualitative results.}
Image examples appear in Appendix~\ref{app:r2c-ablations}
(Figure~\ref{fig:dino-r2c-results}); \pcoarse{} videos and additional image-pair
results are provided on the project page: \projecturl{}.

\subsection{\methodname{} Ablations and Efficiency}
\label{sec:exp-c2r}

\begin{figure}[!htbp]
\centering
\includegraphics[width=1.0\linewidth]{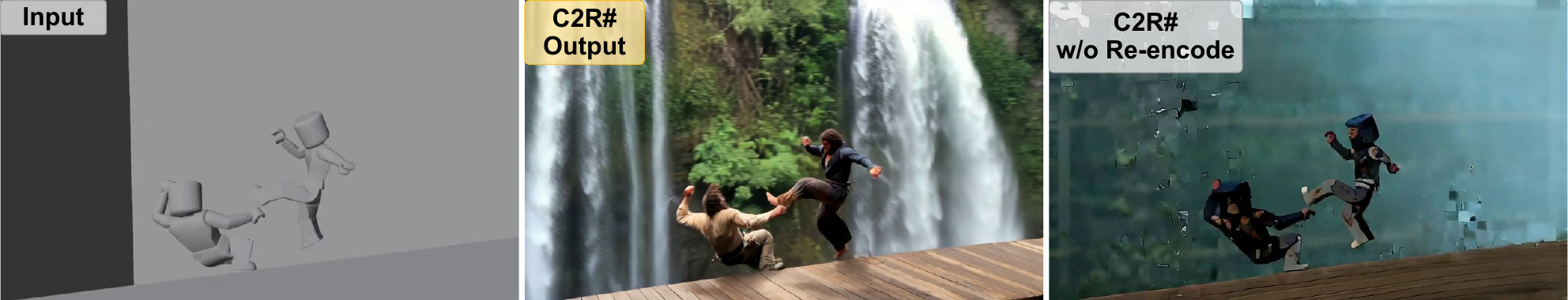}
\caption{\textbf{Effect of the RGB bottleneck during training.}
Matched frames show the control input (left), \methodname{} (center), and a
variant conditioned directly on predicted \dinorname{} features, bypassing
decoding and re-encoding (right). The latter produces visible artifacts.}
\label{fig:native-d2-failure}
\end{figure}

\paragraph{Effect of the RGB bottleneck.}
Without RAE decoding, \method{} is trained directly on \dinor{} features
predicted from real videos. These features may carry hidden realistic
information that is unavailable when actual coarse videos are encoded at
inference, creating a training--inference distribution shift. Decoding through
the RAE acts as an information bottleneck that removes such hidden cues before
the coarse pixels are re-encoded by DINO. Figure~\ref{fig:native-d2-failure}
shows the failure without this bottleneck.
This illustrative variant is excluded from quantitative comparisons.

\paragraph{Efficiency.}
With Wan2.2 frozen, the \method{} trains only 180.54M control and fusion
adapter parameters (0.63\% of the counted model parameters).
Appendix~\ref{app:efficiency} reports parameter comparisons
(Table~\ref{tab:efficiency}) and \dinorname{} training/inference costs
(Table~\ref{tab:r2c-efficiency}).

\subsection{Comparisons}
\label{sec:exp-compare}

\begin{figure}[tp]
\centering
\includegraphics[width=1.0\linewidth]{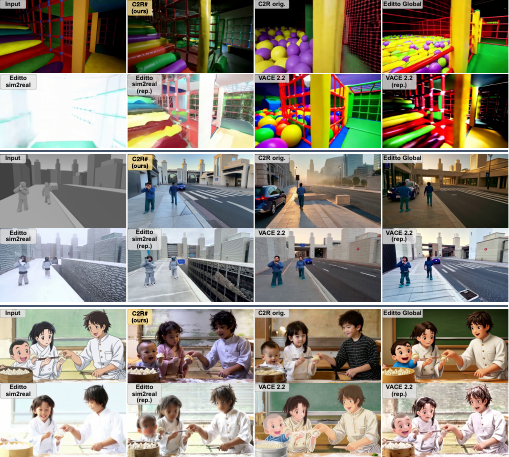}
\caption{
    \textbf{Qualitative comparison with baseline methods.}
    Three groups of two rows show \gameRenders{} (top), \geometryBlockouts{}
    (middle), and \stylizedAnimations{} (bottom).
    ``Rep.'' denotes our Editto Sim2Real and VACE 2.2 reproductions,
    fine-tuned on the same pseudo-paired videos as \methodname{} (ours).\looseness=-1
}
\label{fig:comp-baselines}

\vspace{1.5em}
\centering
\captionof{table}{Comparison on 300 control videos: 100 \gameRenders{},
100 \geometryBlockouts{}, and 100 \stylizedAnimations{}.
Every score averages 300 valid videos per method.
``Rep.'' denotes Editto Sim2Real and VACE 2.2 fine-tuned on the same
pseudo-paired training videos as \methodname{} (ours).
Bold/underlined values rank first/second; arrows indicate preferred directions.
Metric abbreviations and scales appear in Appendix~\ref{app:metrics}.}
\label{tab:c2r-baseline-metrics}
\begingroup
\setlength{\tabcolsep}{3pt}
\renewcommand{\arraystretch}{1.20}
\resizebox{\linewidth}{!}{%
\begin{tabular}{@{}l*{10}{c}@{}}
\toprule
& \multicolumn{2}{c}{DOVER}
& \multicolumn{3}{c}{VEFX}
& VBench
& \multicolumn{3}{c}{Control alignment}
& VLM \\
\cmidrule(lr){2-3}\cmidrule(lr){4-6}\cmidrule(lr){7-7}
\cmidrule(lr){8-10}\cmidrule(lr){11-11}
Method
& \shortstack{Tech.\\$\uparrow$}
& \shortstack{Aesth.\\$\uparrow$}
& \shortstack{RQ\\$\uparrow$}
& \shortstack{IF\\$\uparrow$}
& \shortstack{EE\\$\uparrow$}
& \shortstack{BG\\$\uparrow$}
& \shortstack{DINO\\$\downarrow$}
& \shortstack{Flow\\$\uparrow$}
& \shortstack{Cam.\\$\downarrow$}
& \shortstack{Realism\\$\uparrow$} \\
\midrule
\vone{} & 0.2647 & \underline{0.8033} & \underline{3.5082} & 2.8467 & 2.3468 & 0.9259 & 0.0226 & 0.6041 & 0.1733 & \textbf{75.50} \\
Editto Global & \underline{0.2648} & 0.7700 & 3.4876 & \underline{2.9035} & 2.3637 & 0.9338 & 0.0189 & 0.6897 & 0.1634 & 66.77 \\
Editto Sim2Real & 0.2288 & 0.7124 & 2.6722 & 1.4108 & 1.6995 & 0.9254 & 0.0196 & 0.5986 & 0.1788 & 52.62 \\
Editto Sim2Real (rep.) & 0.2504 & 0.7589 & 3.1011 & 1.7402 & 2.2450 & 0.9283 & \textbf{0.0111} & 0.7314 & 0.1574 & 58.52 \\
VACE 2.2 & 0.2599 & 0.7768 & 3.4110 & 2.5446 & 2.3182 & \textbf{0.9354} & 0.0135 & \textbf{0.7466} & \textbf{0.1491} & 61.88 \\
VACE 2.2 (rep.) & 0.2487 & 0.7117 & 2.9695 & 2.4918 & \underline{2.5334} & 0.9272 & \underline{0.0114} & \underline{0.7347} & \underline{0.1549} & 51.96 \\
\midrule
\textbf{\methodname{}~(ours)} & \textbf{0.2841} & \textbf{0.8169} & \textbf{3.5679} & \textbf{3.1823} & \textbf{2.5923} & \underline{0.9343} & 0.0147 & 0.6588 & 0.1600 & \underline{74.17} \\
\bottomrule
\end{tabular}%
}
\endgroup

\end{figure}

\paragraph{Protocol.}
We compare the \method{} against the model from the \vonepaper{}, Editto Global, the released Editto
Sim2Real, VACE 2.2, and versions of Editto Sim2Real and VACE 2.2 fine-tuned
on exactly the same pseudo-paired training data as the \method{}.
All methods use the same 300 actual coarse videos and scene-specific text
prompts. Table~\ref{tab:c2r-baseline-metrics}
reports DOVER technical and aesthetic quality~\citep{wu2023dover}, the three
VEFX editing scores~\citep{gao2026vefx} (rendering quality, instruction
following, edit exclusivity), VBench background consistency~\citep{huang2024vbench},
three control-alignment measures (DINOv3 feature displacement~\citep{simeoni2025dinov3},
UFM-based optical-flow agreement~\citep{zhang2025ufm}, and ViPE-based camera-pose error~\citep{huang2025vipe}),
and a realism score judged by Qwen3.8-27B-FP8~\citep{qwen2026qwen38}
on a 0--100 scale.\looseness=-1

\paragraph{Results.}
\methodname{} leads DOVER technical and aesthetic quality and all three VEFX
dimensions, ranking second in background consistency and VLM-judged realism. Relative to
the model from the \vonepaper{}, the \method{} improves nine of the ten
reported metrics, including spatial correspondence, local-flow
agreement, and camera-pose error. The model from the \vonepaper{} retains
the higher realism mean (75.50 versus 74.17).
The \method{} exceeds every external baseline on this realism score,
with Editto Global the closest at 66.77.

The results also show a quality--control tradeoff. The adapted
Editto model achieves the lowest DINO displacement, while default
VACE 2.2 leads local-flow agreement and camera-pose accuracy.
Thus, the quality gains of the \method{} do not imply the strongest
preservation under every control metric. Additional plausible detail and secondary
motion may contribute to source--output disagreement.\looseness=-1

Appendix~\ref{app:seedance-comparison} additionally compares first-attempt
results with the commercial model Seedance 2.5~\citep{bytedance2026seedance25}
using the same control videos and text prompts.

\paragraph{Transfer of pseudo-paired training data.}
Adapting Editto Sim2Real with our generated pairs improves all
ten reported aggregate metrics, including realism from 52.62
to 58.52. The effect on VACE 2.2 is mixed: adaptation improves
DINO spatial correspondence and VEFX edit exclusivity, while
reducing the remaining reported scores. These results support
the utility of pseudo-paired supervision beyond the \method{},
while showing that its benefit depends on the recipient model
and adaptation configuration.

\begin{figure}[!t]
    \centering

    \includegraphics[width=\linewidth]{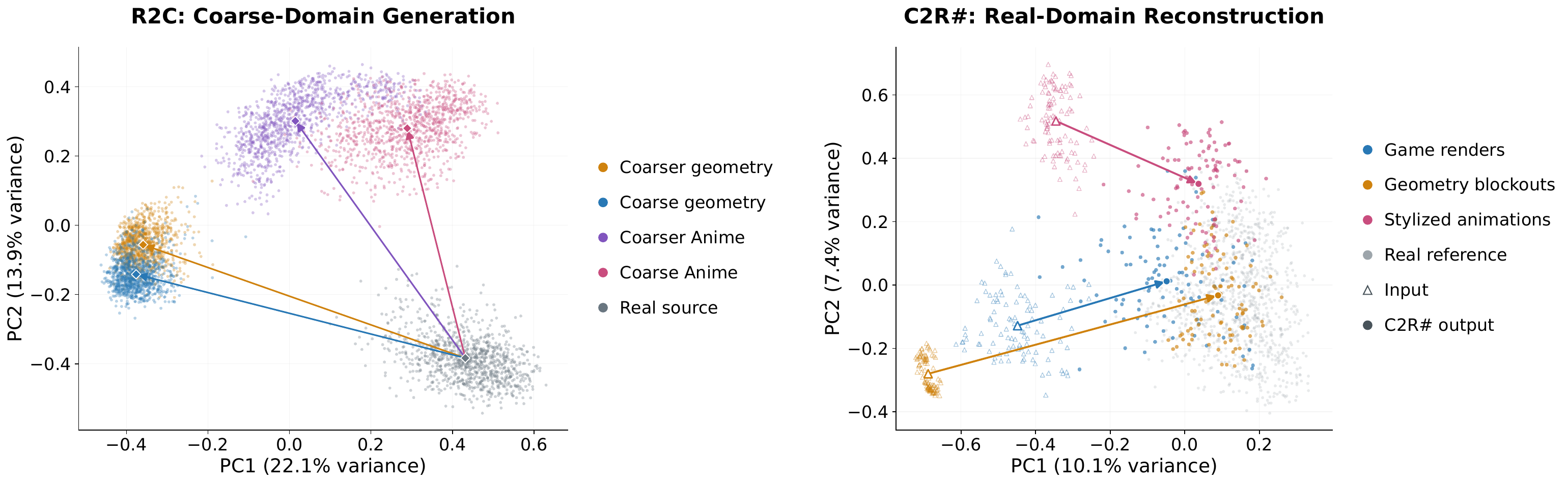}

    \caption{
        \textbf{Domain transfer in CSD-L style-feature space.}
        \textbf{Left:} \dinorname{} maps 1,000 real videos to four distinct
        \pcoarse{} style clusters. \textbf{Right:} \methodname{} maps actual
        coarse videos toward the real-reference distribution.
        Points denote videos; arrows connect input/output centroids.
        PCA is fitted per panel (Appendix~\ref{app:domain-transfer}).
    }
    \label{fig:domain-transfer}
\end{figure}

\paragraph{Domain transfer.}
Figure~\ref{fig:domain-transfer} visualizes both models in the CSD-L
style-feature space~\citep{somepalli2024style}: the \dinor{} moves real videos into four distinct coarse
style clusters, and the \method{} moves each evaluation set of actual coarse
videos toward the distribution of real reference videos.

\paragraph{Reference-image conditioning.}
A separately trained extension uses a reference image to guide appearance
while the coarse video controls layout and motion (details and examples in
Appendix~\ref{app:reference-conditioning}).

\section{Limitations and Discussion}

Our method relies on pseudo-coarse videos generated by \dinor{} to approximate
the coarse inputs encountered at inference. Performance may therefore degrade
when the gap between pseudo-coarse and actual coarse videos becomes large.
Evaluation on actual coarse videos is also challenging because paired realistic
ground truth is unavailable, requiring reference-free metrics and human
judgment.\looseness=-1

\begin{figure}[!htbp]
\centering
\includegraphics[width=0.7\linewidth]{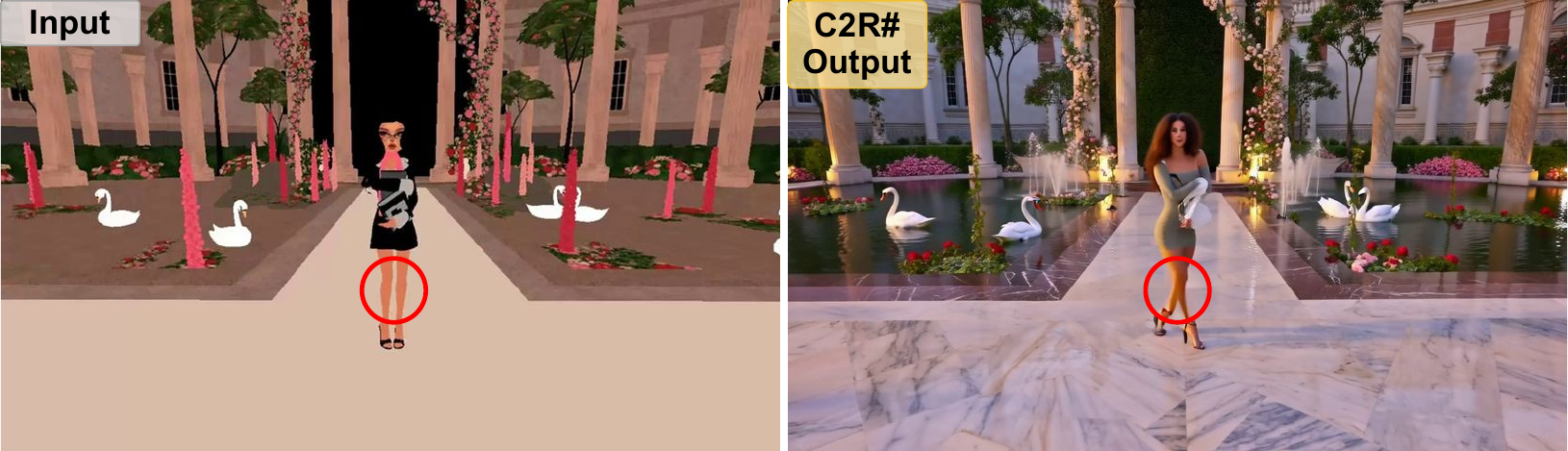}
\caption{
    \textbf{Limitations under abrupt control changes.}
    Input (left) and \methodname{} output (right): abrupt control changes
    can lead to unwanted content and reduced alignment.
}
\label{fig:limitation}
\end{figure}

Our current implementation is constrained by the quality of the
non-commercial backbone model and training data available. The frozen video prior also introduces a trade-off between realism and exact
control alignment. When the input contains abrupt pose or motion changes,
\method{} may adjust the motion pace or synthesize intermediate states rather
than following the discontinuity exactly. This can produce a more natural and
temporally coherent video, but with reduced alignment to the input control
(Figure~\ref{fig:limitation}). Future
work includes closing the gap between pseudo-coarse and actual coarse
features, strengthening reference-image control, and broadening evaluation on
actual coarse videos.

\bibliographystyle{arxiv_preprint}
\bibliography{refs}

\clearpage
\appendix

\section{Implementation Details}
\label{app:implementation}

\subsection{Image-Pair Corpus}
\label{app:pairs}

We use 14,000 aligned (coarse, realistic) image pairs for each of four
modalities: coarser geometry, coarse geometry, coarser anime, and coarse anime.
All realistic source images are generated with
GPT-Image 2~\citep{openai2026gptimage2}. Both source images and coarse targets are
generated at $1280\times720$ (720p). Larger per-modality training sets provided
only marginal improvements, so we use the same 14,000-pair budget for all four
modalities.

\paragraph{Coarse geometry, coarser anime, and coarse anime.}
We use GPT-Image 2 in image-editing mode to create these three types of coarse
target from the realistic source images. Modality-specific instructions request
the desired geometric simplification or anime stylization while preserving the
source subjects, scene layout, poses, and camera viewpoint.

\paragraph{Coarser geometry.}
We reconstruct and simplify a 3D scene from each generated realistic source
image, processing people and background separately. For images containing
people, RF-DETR 2XL~\citep{robinson2026rfdetr} detects them,
SAM 3.1~\citep{meta2026sam31} extracts their masks, and
SAM 3D Body~\citep{yang2026sam3dbody} estimates their 3D bodies.
ObjectClear~\citep{zhao2026objectclear} removes the people and inpaints the
background. OneFormer~\citep{jain2023oneformer} separates ground, sky, and
objects, with SAM 3.1 providing object-instance masks. We reconstruct objects
with Pixal3D~\citep{li2026pixal3d}, use the masked
MoGe~\citep{wang2025moge} point cloud to determine their scene depth, and
triangulate its ground points to recover the terrain, excluding sky regions.
We simplify object meshes into primitives or convex parts and retarget the
reconstructed human poses to blocky Roblox R15 characters using a deterministic
MHR-to-Roblox mapping. Finally, we compose the simplified background and
characters and render from the source camera, preserving scene layout,
perspective, and occlusion to obtain the paired coarser-geometry target.

\subsection{\dinorname{} Architecture and Training}
\label{app:r2c-training}

The frozen encoder $E$ is DINOv3-L/16, using affine-normalized final-layer
patch tokens (layer index 23). At $1280\times720$, it produces a
$45\times80\times1024$ feature grid. The \dinor{} has one trainable DINOv3-L
backbone shared across all four modalities and four modality-specific residual
token-MLP heads, each with hidden width 2048 and a zero-initialized output
projection.

We first fine-tune the RAE decoder $D$~\citep{zheng2026diffusion} for
reconstruction at $1280\times720$ (720p), then freeze it during \dinorname{}
training and pseudo-paired video synthesis. It maps the predicted features
back to coarse RGB pixels. We optimize the \dinor{} with
AdamW~\citep{loshchilov2019adamw} using
learning rate $3\times10^{-5}$, weight decay $0.05$, and gradient clipping at
$1.0$. The shared four-modality model is trained for 20,000 iterations.
Our selected model uses LPIPS and
$2\times$ camera augmentation; Section~\ref{sec:exp-bridge} evaluates the
effect of LPIPS and of the augmentation strength.

\subsection{Camera Augmentation}
\label{app:augmentation}

We apply the same aspect-corrected camera homography to each realistic source
image and its paired coarse target, preserving their alignment. The $1\times$
recipe uses retained crop area in $[0.75,1.00]$, translation bounded by
$\pm0.06$ in normalized image coordinates, rotation by $\pm3.55^\circ$,
horizontal and vertical shear by $\pm2.29^\circ$, and perspective coefficients
by $\pm0.041$. The $2\times$ recipe doubles these perturbation bounds and
doubles the maximum removable crop area, giving a retained-area range of
$[0.50,1.00]$. The $0.5\times$ variant halves the perturbation bounds and
retains crop area in $[0.875,1.00]$. All variants use bilinear resampling
without horizontal flips, photometric augmentation, or padding. The multiplier
controls augmentation strength, not the number of training examples.

\subsection{Pseudo-Paired Video Synthesis}
\label{app:pseudo-pairs}

Our training corpus contains 389,854 real videos drawn exclusively from
DL3DV~\citep{ling2024dl3dv}, \texttt{SEKAI\_HQ}, and
\texttt{SEKAI\_DRONE}~\citep{li2026sekai}. We process clips at
$1280\times720$, $16$ fps, using $81$ frames per clip. During training, the
shared \dinor{} processes each frame independently using the selected
modality head, and the frozen RAE decoder produces the corresponding
\pcoarse{} RGB video. The source real video provides its paired training
target. These pseudo-paired videos are synthesized on the fly, with equal
sampling budgets across the four coarse modalities. We apply the same
resize-to-cover and center-crop transform to the \pcoarse{} control and real
target to obtain $832\times480$ inputs.

\subsection{\methodname{} Architecture and Training}
\label{app:c2r-training}

The video backbone is Wan2.2 T2V-A14B, consisting of a $16$-channel Wan2.1 VAE,
a umT5-xxl text encoder, and two $40$-layer, $5120$-wide MoE DiTs with expert
boundary $\sigma=0.875$. We train two control adapters, one for each expert.
Each adapter contains an input projection $1024\!\to\!512$, four residual 3D
convolution blocks with dilations $1$ and $2$ and FiLM timestep conditioning,
causal temporal compression from $81$ input frames to $21$ latent frames, and a
zero-initialized output projection $512\!\to\!5120$. This is followed by a
zero-initialized fusion adapter consisting of AdaLN and a depthwise
convolution.

The \method{} operates at $832\times480$. At this resolution, DINOv3-L/16
produces a $30\times52$ spatial feature grid. The VAE latent has shape
$(16,21,60,104)$ and, after $2\times2$ spatial patchification, the DiT token
grid is $21\times30\times52$, so each DINO spatial cell corresponds directly to
one DiT spatial token.

We train only the control and fusion adapters using the rectified-flow velocity
loss of Section~\ref{sec:c2r} and AdamW with learning rate $10^{-4}$. The
selected run optimizes 180,540,416 trainable adapter parameters for one epoch
over 389,854 eligible records, corresponding to 12,183 optimization steps with
global batch size 32.

\subsection{Inference Settings}
\label{app:inference}

At inference, the \method{} generates $81$ frames at $16$ fps and
$832\times480$ resolution, using random seed $4321$ for every video.
We resize each input control video while preserving its aspect ratio to cover
$832\times480$, then center-crop to this resolution, applying the same spatial
transform to every frame.

We use a shifted-$\sigma$ schedule with shift $12$ and $50$ denoising steps,
with adaptive projected guidance (APG)~\citep{sadat2025eliminating} at scale
$10$, $\eta=0$, and momentum $-0.5$, adjusting the guidance radius separately
for each evaluation set following that work. Each step uses two forward
passes with the same coarse control: one with the scene prompt and one with
the Wan2.2 default negative prompt. Guidance is applied to the denoised
predictions, then converted back to velocity for the Euler update.

\section{Evaluation Protocol}
\label{app:evaluation}

\subsection{Evaluation Sets}
\label{app:bundles}

We evaluate on three sets of 100 coarse videos each, totaling 300 control
videos. These sets cover \gameRenders{}, \geometryBlockouts{}, and
\stylizedAnimations{} and are used for all compared methods.

\paragraph{Sources of \gameRenders{}.}
The 100 videos are recordings of renderings from Roblox games.

\paragraph{Construction of \geometryBlockouts{}.}
We use Roblox Studio to prototype scenes with an agent. This workflow allowed
us to create all 100 evaluation scenes in under two hours in total and render
them as coarse control videos.

\paragraph{Generation of \stylizedAnimations{}.}
We generate all 100 videos with Wan 2.2 T2V~\citep{wan2025wan}, using text
prompts to produce anime-style animations.

Table~\ref{tab:c2r-baseline-metrics} reports aggregate metrics across all 300
videos, with equal weight per video.

\subsection{Compared Methods}
\label{app:baselines}

The adapted VACE 2.2 and Editto Sim2Real baselines are fine-tuned on exactly
the same pseudo-paired training data as the \method{}: the same 389,854 real
videos from DL3DV, \texttt{SEKAI\_HQ}, and \texttt{SEKAI\_DRONE}, paired with
the corresponding \pcoarse{} controls synthesized by our shared \dinor{}.
All three methods use the same four coarse modalities with equal sampling
budgets and the same $832\times480$, 81-frame, 16-fps media format.

\subsection{Metric Definitions}
\label{app:metrics}

\paragraph{\dinorname{} image metrics (Table~\ref{tab:r2c-image-ablation}).}
\label{app:r2c-image-metrics}
Each realistic input image has a paired coarse target. All eight metrics
compare the prediction with this target and measure how faithfully the model
reproduces the intended coarse image. The model operates at $1280\times720$.
Latent MSE uses the native $45\times80\times1024$ feature grid before decoding;
the other metrics evaluate decoded images clipped to $[0,1]$ and resized to
$256\times256$ with bicubic interpolation. Table entries are means over the
1,000 held-out image pairs.

\paragraph{Latent MSE ($\downarrow$).}
Mean squared error measures the average squared difference between the
predicted feature grid and the target grid produced by the frozen encoder.
It tests whether the model predicts the intended coarse representation before
the RAE decoder turns it into pixels. Zero means identical feature grids;
lower values indicate more accurate feature reconstruction. Its scale differs
from pixel-space MSE, so the two columns should not be compared numerically.

\paragraph{RGB MSE ($\downarrow$).}
We average the squared differences between corresponding output and target
pixel values across all three color channels. This measures pixel-level color
and appearance error, giving larger mistakes more weight because differences
are squared. Zero means an exact pixel match. Small shifts in otherwise
similar shapes can still produce substantial pixel error.

\paragraph{RGB L1 ($\downarrow$).}
We average absolute, rather than squared, differences between corresponding
RGB values. This also measures pixel-level reconstruction error, but gives
less emphasis to a few large mistakes than MSE. With RGB values in $[0,1]$,
an L1 value of $0.1$ means an average absolute channel-value difference of
$0.1$; it is not a percentage of incorrectly reconstructed pixels.

\paragraph{LPIPS ($\downarrow$).}
Learned Perceptual Image Patch Similarity~\citep{zhang2018unreasonable}
compares output and target using features from a frozen perceptual network.
It measures differences in visual appearance, including texture and structure,
that exact pixel comparisons may describe poorly. Despite its name, the
reported quantity is a distance: lower values indicate greater perceptual
similarity. It measures resemblance to the coarse target, not photorealism.

\paragraph{DINO patch cosine ($\uparrow$).}
We extract DINOv3 patch features~\citep{simeoni2025dinov3} from the decoded output and target, compute
cosine similarity at corresponding spatial locations, and average across
patches. This measures local agreement in scene content and structure. A value
closer to one indicates more similar feature directions, independently of
their magnitudes. It can remain high despite some color or texture differences
and does not replace pixel-level fidelity measures.

\paragraph{Depth RMSE ($\downarrow$).}
We estimate relative-depth maps for the output and target using Depth Anything
V2~\citep{yang2024depthanythingv2}, align their scale and offset for each image pair, and compute the root
mean squared difference. This measures agreement in the inferred scene
geometry after removing arbitrary depth scale and offset. Lower values
indicate closer geometric agreement. The score uses estimated relative depth,
not measured distances, and is not an error in meters.

\paragraph{SSIM ($\uparrow$).}
Structural Similarity~\citep{wang2004ssim} compares local brightness, contrast, and spatial
structure between output and target. Our implementation uses luminance
(BT.601 luma) and an $11\times11$ Gaussian window. Higher values, approaching
one, indicate more similar local image structure. Because it operates on
luminance, it does not directly measure color agreement.

\paragraph{Boundary F1 ($\uparrow$).}
We extract Sobel edges from output and target, retain responses above $0.08$,
and match edges within a two-pixel tolerance. The F1 score combines precision
(how many predicted edges match target edges) and recall (how many target
edges are recovered). Extra or missing outlines reduce the score; higher
values indicate better boundary reconstruction. This compares outlines with
the paired target, whereas Table~\ref{tab:r2c-video-ablation}'s boundary
stability measures their consistency between neighboring video frames.

\paragraph{\dinorname{} video metrics (Table~\ref{tab:r2c-video-ablation}).}
\label{app:r2c-video-metrics}
There are no ground-truth coarse target videos. We compare the temporal
behavior of the \pcoarse{} output with the original real video, using 81
frames per clip resized to $256\times256$.
RAFT-Small~\citep{teed2020raft} estimates motion between neighboring source
frames in both temporal directions. We use these correspondences to align
neighboring source and output frames before comparing their appearance,
compensating for motion present in the source. We retain only in-bounds
correspondences for which following the forward flow and then the backward
flow returns close to the starting pixel. This consistency mask reduces the
influence of occlusions and unreliable motion estimates. Table entries are
means of the corresponding per-video scores.

\paragraph{Perceptual flicker (relative, $\downarrow$).}
We measure LPIPS between neighboring output frames after aligning them using
source motion. For each adjacent-frame comparison, this perceptual error is
divided by the corresponding motion-aligned source-video error. We take the
median of these ratios within each video to summarize typical temporal
behavior, then average across videos. A value of one means source-level error;
values below or above one indicate less or more error under this normalization.
This is a temporal error ratio, not a reconstruction-accuracy percentage.

\paragraph{Smoothed-frame flicker (relative, $\downarrow$).}
We spatially blur frames and measure their absolute RGB difference (L1) after
motion alignment. Blurring reduces sensitivity to fine texture and emphasizes
changes in broad colors and shapes. As for perceptual flicker, each output
error is divided by its corresponding source error, and per-video medians are
averaged across videos. Both flicker columns therefore summarize typical
behavior using the same aggregation rule. Lower values indicate less
motion-aligned appearance instability; smoothing can itself reduce these
errors, so they should be interpreted alongside the motion and content metrics.

\paragraph{Motion error (EPE, $\downarrow$).}
Endpoint error measures the mean distance between optical-flow vectors
estimated independently from the output and source videos. Differences in
either direction or magnitude increase the error. Lower values indicate
closer agreement with source motion. Both motion fields are estimated from
images; the source flow is a reference, not ground-truth motion annotation.

\paragraph{Motion magnitude ratio ($\approx 1$).}
For each video, we divide mean output flow magnitude by mean source flow
magnitude. A ratio below one indicates attenuated motion, while a ratio above
one indicates amplified motion; one is preferred. For example, a value of
$1.0797$ indicates approximately $8\%$ greater estimated motion magnitude.
This metric measures the amount of motion, not its direction, and complements
motion error.

\paragraph{Boundary stability (Dice, $\uparrow$).}
We extract edge maps from neighboring output frames, align them using source
motion, and measure their overlap with a soft Dice score. Higher overlap
indicates that outlines persist more consistently across frames. This compares
the output with its temporal neighbors, not with ground-truth coarse boundaries.

\paragraph{Content consistency (DINO, $\uparrow$).}
We spatially average DINOv3 features within each output frame and compute mean
cosine similarity between consecutive frame descriptors. Higher similarity
indicates more stable overall scene and subject content. Spatial averaging can
hide local flicker, and an overly static output can also score highly, so this
metric should be read together with the motion-preservation measures.

\paragraph{\methodname{} metrics (Table~\ref{tab:c2r-baseline-metrics}).}
\paragraph{Perceptual quality and editing assessment.}
Table~\ref{tab:c2r-baseline-metrics} separates output quality,
editing success, control alignment, and realism.
DOVER~\citep{wu2023dover} technical quality (Tech.) measures technical visual quality,
while its aesthetic branch (Aesth.) assesses aesthetic appeal.
VEFX-Reward-32B~\citep{gao2026vefx} assesses the control video, editing instruction,
and generated result on three 1--4 scales: rendering quality
(RQ), including visual clarity, temporal stability, and artifacts;
instruction following (IF), measuring execution of the requested edit;
and edit exclusivity (EE), assessing unintended changes outside
the requested transformation. Because coarse-to-real generation
intentionally changes appearance throughout the scene, EE is
interpreted relative to the editing request, rather than as a
direct measure of spatial or motion alignment.

\paragraph{Temporal consistency and control alignment.}
VBench~\citep{huang2024vbench} background consistency (BG) measures the temporal consistency
of generated background appearance. It does not directly measure
agreement with source motion.
The DINO metric measures normalized spatial displacement of matched
DINOv3 features~\citep{simeoni2025dinov3} between synchronized control and output frames;
lower values indicate closer spatial correspondence among successfully
matched features.
Flow denotes local optical-flow agreement between the control
and generated videos, estimated with UFM~\citep{zhang2025ufm}
(UFM-Base-DINOv2G-init), with covisibility handling and scale
normalization. Cam. denotes relative camera-pose error computed
from ViPE-estimated trajectories~\citep{huang2025vipe}, measuring camera preservation
separately from local object motion. It is an error under the
evaluation protocol, not an absolute distance in meters or an
angle in degrees.

\paragraph{VLM-judged realism.}
We use Qwen3.8-27B-FP8~\citep{qwen2026qwen38} to assess generated videos sampled at
8 frames per second. The judge assigns four scores on a 1--5
scale: appearance, structural plausibility, temporal stability,
and motion and physics. For video $i$, we compute
\begin{equation}
R_i = 25\left(\frac{
s_{i,\mathrm{app}}+s_{i,\mathrm{str}}+
s_{i,\mathrm{temp}}+s_{i,\mathrm{motion}}}{4}-1\right).
\label{eq:vlm-realism-score}
\end{equation}
This maps the mean rating to a 0--100 scale. All evaluation sets use
photorealistic live-action footage as the target domain.
The scene prompt provides context, but this assessment scores
visible realism rather than prompt adherence or source-control
similarity. In particular, plausible output motion does not
necessarily imply preservation of the input trajectory.

\subsection{Domain-Transfer Visualization}
\label{app:domain-transfer}

For Figure~\ref{fig:domain-transfer}, each video is represented by averaging
unit-normalized CSD-L style descriptors~\citep{somepalli2024style} from nine frames at 480p and
normalizing the resulting vector. The left panel contains the same 1,000 real
videos and their four \pcoarse{} versions; the right panel contains the 300
actual coarse control videos, their \methodname{} outputs, and 1,000 unmatched
real reference videos. PCA is fitted independently for each panel with equal
weight on every video; axes show the first two principal-component scores, and
percentages give the variance explained by each component within that panel.

\clearpage
\subsection{Additional \dinorname{} Results}
\label{app:r2c-ablations}

\paragraph{Image ablations.}
Table~\ref{tab:r2c-image-ablation} compares the six training recipes on 1,000
held-out \modGeomCoarse{} image pairs. The $2\times$ augmentation recipe with
LPIPS achieves the best latent MSE, RGB L1, LPIPS, DINO patch cosine similarity,
and depth RMSE. It does not optimize every reconstruction criterion:
$1\times$ augmentation without LPIPS achieves the best RGB MSE and SSIM,
while training without augmentation with LPIPS achieves the best boundary F1.
Metric definitions are given in Appendix~\ref{app:r2c-image-metrics}.

\begin{table}[!htbp]
\centering
\caption{\textbf{\dinorname{} design ablations, image quality.}
Evaluated on 1,000 held-out images in the \modGeomCoarse{} modality, comparing
predictions with paired coarse targets. Latent MSE measures feature
reconstruction; RGB MSE and L1 measure pixel errors; LPIPS and DINO patch
cosine measure perceptual and local content agreement; depth RMSE, SSIM, and
boundary F1 assess geometry, local image structure, and outlines.
Latent MSE uses the native feature grid; other metrics use $256\times256$
images. Values are means across images. Bold and underlining mark the best
and second-best values. Definitions are given in
Appendix~\ref{app:r2c-image-metrics}.}
\label{tab:r2c-image-ablation}

\begingroup
\setlength{\tabcolsep}{4pt}
\renewcommand{\arraystretch}{1.15}
\resizebox{\linewidth}{!}{%
\begin{tabular}{@{}lcccccccc@{}}
\toprule
\dinorname{} Recipe
& \shortstack{Latent\\MSE\\$\downarrow$}
& \shortstack{RGB\\MSE\\$\downarrow$}
& \shortstack{RGB\\L1\\$\downarrow$}
& \shortstack{LPIPS\\$\downarrow$}
& \shortstack{DINO patch\\cosine\\$\uparrow$}
& \shortstack{Depth\\RMSE\\$\downarrow$}
& \shortstack{SSIM\\$\uparrow$}
& \shortstack{Boundary\\F1\\$\uparrow$} \\
\midrule
No Aug, no LPIPS
& 0.4443 & 0.0217 & 0.1138 & 0.3980
& 0.9055 & 0.4537 & 0.6225 & 0.3995 \\

No Aug, w/ LPIPS
& 0.4406 & 0.0218 & 0.1137 & 0.3707
& 0.9121 & 0.4447 & 0.6122 & \textbf{0.5150} \\

$0.5\times$ Aug, w/ LPIPS
& 0.4296 & 0.0214 & 0.1129 & 0.3700
& \underline{0.9128} & 0.4380
& 0.6236 & \underline{0.4654} \\

$1\times$ Aug, no LPIPS
& \underline{0.4276} & \textbf{0.0211}
& \underline{0.1123} & 0.4127
& 0.9020 & 0.4483 & \textbf{0.6359} & 0.2889 \\

$1\times$ Aug, w/ LPIPS
& 0.4280 & 0.0212 & 0.1126 & \underline{0.3682}
& 0.9127 & \underline{0.4357} & 0.6247 & 0.4644 \\

\addlinespace[2pt]
\textbf{$2\times$ Aug, w/ LPIPS (ours)}
& \textbf{0.4237} & \underline{0.0212}
& \textbf{0.1122} & \textbf{0.3659}
& \textbf{0.9136} & \textbf{0.4352}
& \underline{0.6282} & 0.4573 \\
\bottomrule
\end{tabular}%
}
\endgroup
\end{table}

\begin{figure}[!htbp]
\centering
\includegraphics[width=\linewidth]{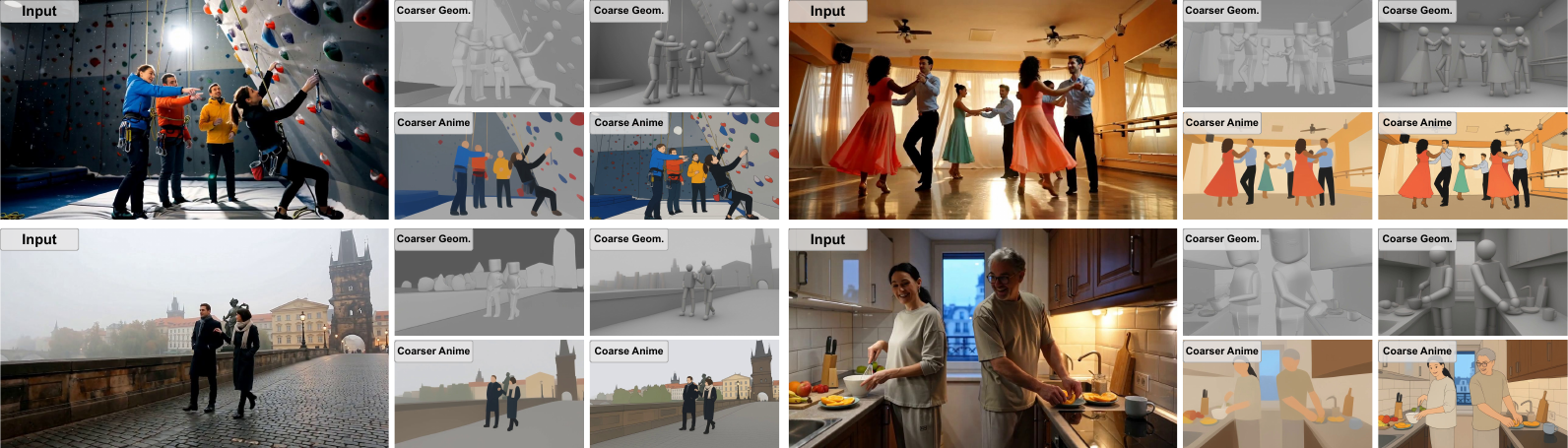}
\caption{Qualitative \dinorname{} results. Each example shows a realistic image
and decoded coarse predictions for \modGeomCoarser{}, \modGeomCoarse{},
\modAnimeCoarser{}, and \modAnimeCoarse{}. The predictions illustrate
modality-dependent appearance transfer while retaining scene structure.}
\label{fig:dino-r2c-results}
\end{figure}

\paragraph{Qualitative results.}
Figure~\ref{fig:dino-r2c-results} shows realistic source images and decoded
coarse predictions across the four modalities. \Pcoarse{} videos, obtained by
applying the \dinor{} to each frame of real clips, are provided on the
project page (\projecturl{}), together with additional image-pair results.

\clearpage
\subsection{Efficiency and Computational Cost}
\label{app:efficiency}

\paragraph{Training parameters.}
Because the Wan2.2 backbone stays frozen, the \method{} trains only its
control and fusion adapters. Table~\ref{tab:efficiency} distinguishes total
model capacity from the parameters optimized during adaptation.
The training procedure in the \vonepaper{} fine-tunes the diffusion backbone
in Stage I, then freezes it and trains the added control and fusion modules
in Stage II; totals reflect the modules present in each stage.
Editto fine-tunes LoRA weights on an existing frozen VACE control branch,
which remains included in its total parameter count. The VACE 2.2 trainable
count follows its documented control-module training configuration.
The \method{} optimizes 180.54M adapter parameters, or 0.63\% of the counted
model parameters. For VACE 2.2 and the \methodname{}, counts sum both denoising
experts, although only one expert is active at each denoising step.

\begin{table}[!htbp]
\centering
\caption{Training parameters.
Totals include the diffusion backbone and control/fusion modules,
including unmerged LoRA where applicable, but exclude frozen
external text and feature encoders, the VAE, and separately
trained \dinorname{} models.
Counts include both denoising experts for VACE 2.2 and the \methodname{}.
B and M denote billions and millions of parameters.}
\label{tab:efficiency}
\begingroup
\setlength{\tabcolsep}{5pt}
\renewcommand{\arraystretch}{1.15}
\small
\begin{tabular}{@{}lccc@{}}
\toprule
Method / training stage
& \shortstack{Total\\params}
& \shortstack{Trainable\\params}
& \shortstack{Trainable\\(\%)} \\
\midrule
\vone{}: Stage I
& 14.288B & 14.288B & 100.00 \\
\vone{}: Stage II
& 14.333B & 44.33M & 0.31 \\
\midrule
Editto Global
& 17.460B & 122.68M & 0.70 \\
Editto Sim2Real
& 17.460B & 122.68M & 0.70 \\
VACE 2.2
& 34.675B & 6.098B & 17.59 \\
\midrule
\methodname{} (ours)
& 28.758B & 180.54M & 0.63 \\
\bottomrule
\end{tabular}
\endgroup
\end{table}

\paragraph{\dinorname{} efficiency.}
Table~\ref{tab:r2c-efficiency} reports the shared four-head model's parameter
count, training hardware and cost, and inference performance at native 720p.
The inference timings include the projector and frozen RAE decoder and exclude
model loading, transfers, video codecs, and storage I/O.

\begin{table}[!htbp]
\centering
\caption{Model size and efficiency for
a shared-backbone, four-head \dinorname{} architecture. Inference uses BF16 at $1280\times720$ and includes the
projector and frozen RAE decoder, excluding model loading,
transfers, video codecs, and storage I/O.}
\label{tab:r2c-efficiency}
\small
\begin{tabular}{lr}
\toprule
Quantity & Value \\
\midrule
\multicolumn{2}{l}{\textit{Model params}} \\
Shared backbone + four heads & $319.92$M trainable \\
\midrule
\multicolumn{2}{l}{\textit{Training efficiency}} \\
Hardware & $3 \times 8$ B200 GPUs \\
Training iterations & 20,000 \\
Global batch size & 192 \\
Recorded training time & 4h 02m 58s \\
Derived training cost & 97.19 GPU-hours \\
Median training step time & 0.703 s \\
\midrule
\multicolumn{2}{l}{\textit{Inference efficiency}} \\
Hardware & 1 B200 GPU \\
Image latency, batch 1 & 34.779 ms \\
Image latency, batch 8 & 199.303 ms/batch \\
Video length / frame chunk size & 81 / 8 \\
Median video latency & 2.029 s \\
Video throughput & 39.92 frames/s \\
Peak allocated GPU memory, video & 4.820 GiB \\
\bottomrule
\end{tabular}
\end{table}

\subsection{Reference-Conditioned Generation}
\label{app:reference-conditioning}
\label{sec:reference-conditioning}

The \method{} can additionally accept a reference image alongside the control video and
text prompt, using separately trained reference-conditioned weights. The extension
retains the control and fusion adapters for the control video and the original text
conditioning. We encode the reference image with the frozen Wan VAE and append its
tokens after the generation tokens. A temporal rotary-position offset of 100 separates
the reference from the generated sequence, with reference-specific adaptive layer
normalization for each expert. Rank-16 LoRA updates~\citep{hu2022lora}
(scaling parameter 16) adapt the
query, key, value, and output projections of self-attention in all 40 blocks of both
Wan experts. The original backbone weights remain frozen; the reference extension
trains the adapters and LoRA parameters. The coarse video still controls layout and motion, while the reference image
specifies additional appearance cues.
Because this use case receives an extra input, we present it separately from the
standard control-video-and-text comparisons.

\begin{figure}[!htbp]
\centering
\includegraphics[width=1.0\linewidth]{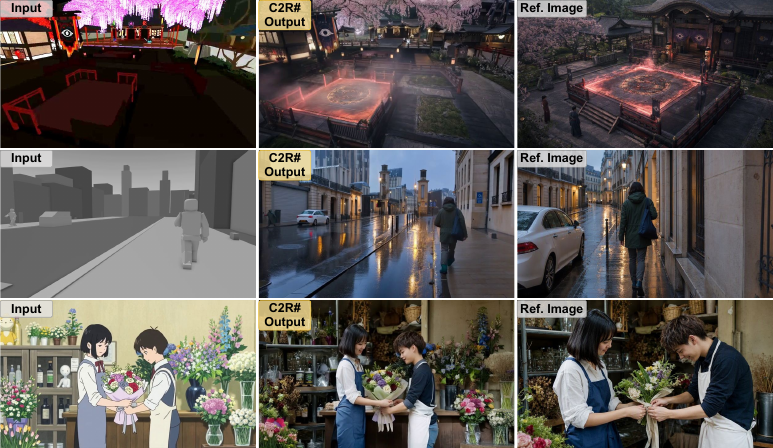}
\caption{The \method{} with a reference image. Each example includes a control video, the reference image
used at inference, a text prompt, and the generated result. Examples cover
\gameRenders{}, \geometryBlockouts{}, and \stylizedAnimations{}. This extension uses separately
trained weights and an additional image input, and is evaluated separately from
the standard control-video-and-text setting.}
\label{fig:reference-conditioning}
\end{figure}

\clearpage
\subsection{Qualitative Comparison with Seedance 2.5}
\label{app:seedance-comparison}

\begin{figure}[!htbp]
\centering
\includegraphics[width=\linewidth]{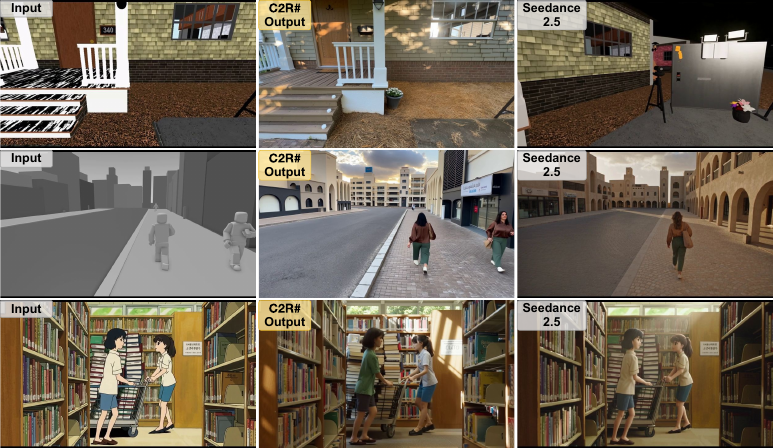}
\caption{\textbf{First-attempt comparison with Seedance 2.5.}
Columns show the input control, \methodname{} output, and Seedance 2.5 output.
Rows show examples of \gameRenders{}, \geometryBlockouts{}, and
\stylizedAnimations{}, respectively. Both methods receive the same control
video and text prompt, and the displayed results are from the first generation
attempt. In these examples, the \method{} produces a more photorealistic
appearance, while Seedance 2.5 retains more of the coarse input style or
changes the scene layout.}
\label{fig:seedance-comparison}
\end{figure}

We compare the \method{} with Seedance 2.5~\citep{bytedance2026seedance25},
a closed-source commercial video model, using the same control videos and
text prompts on inputs drawn from our three evaluation sets. In our trials,
when realistic domain transfer succeeds, Seedance 2.5 produces highly realistic
results comparable in photorealism to the \method{}. However, for this
coarse-to-real task, we observed that it is prone to losing alignment with the
input control video and can fail to produce the photorealistic appearance
requested by the prompt. Figure~\ref{fig:seedance-comparison} shows
first-attempt examples of these failures: Seedance 2.5 changes the scene layout
or retains visibly stylized or game-like elements, while the \method{}
more faithfully preserves the input structure and achieves the requested
photorealistic appearance. We also found commercial editing expensive per
video, making retries costly.

\end{document}